\pdfoutput=1

\documentclass[11pt]{article}

\usepackage{ACL2023}

\usepackage{times}
\usepackage{latexsym}
\usepackage{graphicx}

\usepackage[T1]{fontenc}

\usepackage[utf8]{inputenc}

\usepackage{microtype}

\usepackage{booktabs}
\usepackage[normalem]{ulem}
\useunder{\uline}{\ul}{}
\usepackage{amsfonts}
\usepackage{amsmath}

\usepackage{inconsolata}

\title{Retrieval-Augmented Classification with Decoupled Representation}

\author{Xinnian Liang\textsuperscript{1}\footnotemark[1], Shuangzhi Wu\textsuperscript{2}, Hui Huang\textsuperscript{3}, Jiaqi Bai\textsuperscript{1}, Chao Bian\textsuperscript{2}, Zhoujun Li\textsuperscript{1}\footnotemark[2]\\ 
\textsuperscript{1}State Key Lab of Software Development Environment, Beihang University, Beijing, China \\ 
\textsuperscript{2}Lark Platform Engineering-AI, Beijing, China\\ 
\textsuperscript{3}Faculty of Computing, Harbin Institute of Technology, Harbin, China\\
\texttt{\{xnliang,bjq,lzj\}@buaa.edu.cn,\{wufurui,zhangchaoyue.0,huanghui.hit\}@bytedance.com}
}

\begin{document}
\maketitle
\renewcommand{\thefootnote}{\fnsymbol{footnote}} 
\footnotetext[1]{Contribution during internship at ByteDance Inc.} 
\footnotetext[2]{Corresponding Authors.} 
\renewcommand{\thefootnote}{\arabic{footnote}} 

\begin{abstract}

Retrieval augmented methods have shown promising results in various classification tasks. However, existing methods focus on retrieving extra context to enrich the input, which is noise sensitive and non-expandable. In this paper, following this line, we propose a $k$-nearest-neighbor (KNN) -based method for retrieval augmented classifications, which interpolates the predicted label distribution with retrieved instances' label distributions. Different from the standard KNN process, we propose a decoupling mechanism as we find that shared representation for classification and retrieval hurts performance and leads to training instability. We evaluate our method on a wide range of classification datasets. Experimental results demonstrate the effectiveness and robustness of our proposed method.  We also conduct extra experiments to analyze the contributions of different components in our model.\footnote{\url{https://github.com/xnliang98/knn-cls-w-decoupling}}

\end{abstract}

\section{Introduction}
Retrieval augmented methods have been widely used in many Natural Language Processing (NLP) tasks, such as question answering~\cite{yang-etal-2021-neural-retrieval,mao-etal-2021-generation}, semantic parsing~\cite{pasupat-etal-2021-controllable,DBLP:journals/csur/DongLGCLSY23}, code generation~\cite{lu-etal-2022-reacc}, classification~\cite{10.1007/978-3-031-16210-7_56,gur-etal-2021-cross-modal}, etc. Existing retrieval-augmented models attached several retrieved texts as knowledge to the original inputs to improve performance. However, they required an extra corpus and a single retrieval model to obtain the knowledgeable context, which makes them non-expandable. In addition, retrieved text sometimes brings noise into the original input. 

Recently, $k$-nearest-neighbor (KNN) -based methods were successfully applied in language modeling~\cite{khandelwal20knnlm}, machine translation~\cite{khandelwal2021knnmt,zheng-etal-2021-adaptive}, and multi-label classification~\cite{su-etal-2022-contrastive}. 
KNN-based methods first build a datastore over the labelled dataset. The datastore is a series of key-value pairs, where the key is the representation of each instance and the value is the label. 
Then, during the prediction stage, they use the representation of the input instance to retrieve $k$ nearest pairs from the datastore. Finally, the labels of similar instances are used to interpolate the predicted label distribution. 
From the previous process, we can see that the KNN-based method does not introduce noise into the model inputs and does not need to train a single retrieval model.

\begin{figure*}
    \centering
    \includegraphics[width=0.7\textwidth]{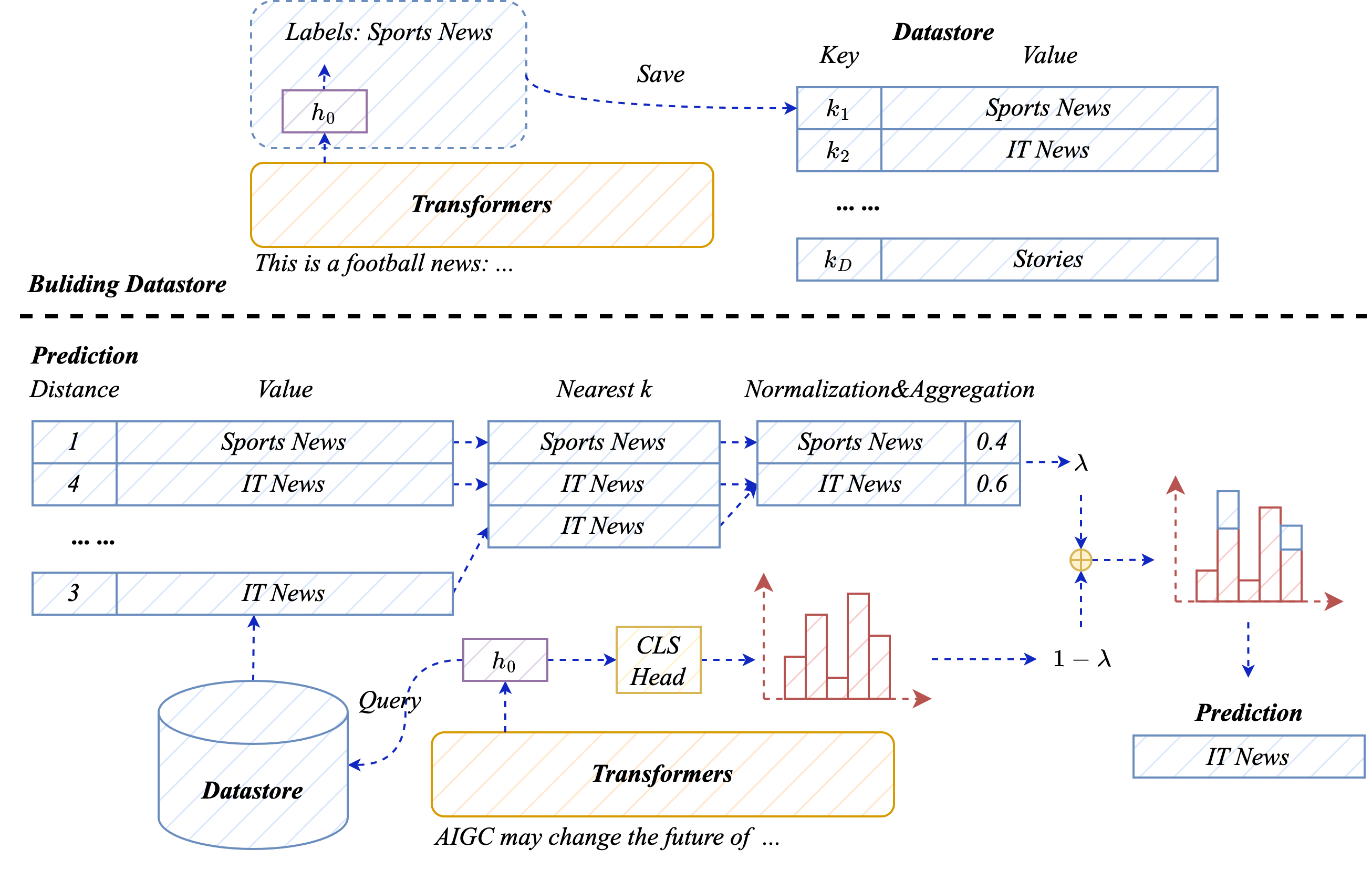}
    \caption{The illustration of how to build the datastore and predict the final label with KNN retrieval. Notes that $k_i$ denotes the query and key vector for retrieval.}
    \label{fig:main}
\end{figure*}

Following this line, in this paper, we propose a $k$-nearest-neighbor (KNN) -based method for retrieval augmented classifications, which retrieves $k$ nearest instances from the training data~\cite{wang-etal-2022-training} and interpolates the predicted label distribution with the retrieved instances' labels. Different from the existing KNN process, we propose a simple yet effective decoupling mechanism for KNN to tackle the issue found during experiments, where shared representation for classification and retrieval hurts performance and training stability.
Specifically, we first fine-tune one pre-trained language model (PLM) with our proposed decoupling mechanism on one specific classification dataset. 
The decoupling mechanism consists of one decoupling layer and one training loss. The former decouples the retrieval representation from the PLM output, and the latter guide the retrieval representation to learn the distance between different instances.
After that, we employ learned retrieval representation as the key to build the datastore. During prediction, the predicted label distribution is from the output classification representation and the retrieval representation is used to retrieve $k$ nearest instances from the datastore. Finally, the model outputs the label with predicted label distribution and retrieved label distribution.


We evaluate our methods on a wide range of classification datasets, including six Chinese and six English datasets. Experimental results demonstrate the effectiveness of our proposed KNN-based classification method and decoupling module. We also conduct extra experiments to analyze the contributions of different components.

\section{Methodology}
In this section, we first introduce the working progress of the $k$-nearest-neighbor (KNN) -based classification, which is shown in Figure~\ref{fig:main}. After that, we introduce the decoupling mechanism.

\subsection{Nearest Neighbor based Classification}
We first need one classification model to provide high-quality instance representation and predict the label distribution of input instances.
Then, as shown in Figure~\ref{fig:main}, we employ this model to build the datastore, which contains key-value pairs. Finally, during prediction, the model retrieves $k$ instances from the datastore to involve the predicted label distribution.



\paragraph{PLM Fine-tuning} To obtain a proper model, we first fine-tune one PLM on the training set $\mathcal{D}=\{s_i, l_i\}_{i=1}^N$, where $N$ is the number of instances, $s_i$ is the input sentence, and $l_i$ is the label. 
Formally, given an input sentence $s_i$, we encode it with the PLM $\mathcal{M}_{\theta}$ and obtain the representations $\{h_0, \dots, h_{L}\} = \mathcal{M}_{\theta}(s_i)$.
Then, we choose the representation $h_0$ as the input for the classification head to predict the probability distribution of labels as follows:
\begin{equation}
    \mathcal{P}_{CLS}(y|s_i) = \mathtt{Softmax}(\sigma(W^o\cdot h_0))
\end{equation} where $W^o \in \mathcal{R}^{emb\_size\times num\_labels}$, $\sigma(\cdot)$ is activation function. 
Finally, the cross entropy loss $\mathcal{L}_{CE}$ based on golden label $l_i$ and predicted label distribution $\mathcal{P}_{CLS}(y|s_i)$ is computed to update the model parameters $\theta$.

\paragraph{Building Datastore} The progress of building a datastore is shown in the top part of Figure~\ref{fig:main}. 
The datastore contains a series of key-value pairs $(k_i, v_i) \in (\mathcal{K}, \mathcal{V})$, where the key is the instance representation $h_0$ from $\mathcal{M}_{\theta}(s_i)$ and value is the golden reference $l_i$. Specifically, we build one specific datastore for each classification dataset over its training data.

\paragraph{Prediction} During prediction, given an input $s_i$ from test set, the model $\mathcal{M}_{\theta}$ encode it as $h_0$. Then, the $h_0$ is used as query $h_q$ to search the nearest $k$ neighbors $(k_j, l_j) \in (\mathcal{K}, \mathcal{V})$ according to squared-$L^2$ distance, $d$.
We employ the FAISS~\cite{johnson2019billion}, which is a library\footnote{https://github.com/facebookresearch/faiss} for fast nearest neighbour search, to obtain the neighbours.

As shown in the bottom part of Figure~\ref{fig:main}, the retrieved $k$ neighbours' labels are converted into a probability distribution by applying a softmax to the scaled negative distances and aggregating the probability over the same label items. 
The computation is as follows:
\begin{equation}
    \mathcal{P}_{KNN}(y|s_i) \propto \sum_{(k_j, l_j)\in \mathcal{N}}\mathbb{I}_{y=l_i}\mathtt{exp}(\frac{-d(k_j,h_q)}{T})
\end{equation} where $d(\cdot)$ is squared-$L^2$ distance, $T$ is a temperature to scale the distance, which is from~\cite{khandelwal2021knnmt} and we empirically set it as 10.
Finally, we interpolate two distributions as follows:
\begin{equation}
    \mathcal{P}(y|s_i) = \lambda \mathcal{P}_{KNN}(y|s_i) + (1-\lambda) \mathcal{P}_{CLS}(y|s_i)
\end{equation} where the $\lambda$ is a hyper-parameter to adjust the influence of KNN retrieved label distribution.

\subsection{Decoupling Mechanism}
Since experiments, we find that if we use the same vector representation for classification and retrieval, the model training is unstable and performance dropped. To tackle this issue, we propose a decoupling mechanism, which contains one decouple layer and one training loss. Precisely, we use a separate representation $r_i$ to decouple the retrieval ability from $h_0$ to ensure the $h_0$ is only used for the label prediction in Equation (1). The separate representation $r_i$ is used as the instance representation and is obtained by a simple MLP layer $r_i=\mathtt{MLP}(h_0)$.
Intuitively, the representation $r_i$ should have the ability to measure the similarity of different instances. Therefore, during training, we force the $r_i$ to be closer to the positive example $r_+$ and farther to the negative example $r_-$ by adding a triplet loss into the training loss as follows: 
\begin{equation}
    \begin{aligned}
        &\mathcal{L} = (1-\beta)\mathcal{L}_{CE} + \beta \mathcal{L}_{DIS} \\
        &\mathcal{L}_{DIS} = \max(d(r_i, r_+)-d(r_i, r_-)+\mu, 0) 
    \end{aligned}
\end{equation}
Where $\mathcal{L}_{CE}$ is the Cross-Entropy loss for label prediction, $\mathcal{L}_{DIS}$ is triplet loss for learning instance representations, and $d(\cdot)$ is squared-$L^2$ distance, which is aligned to the distance metric of the KNN retrieval. The positive/negative examples are selected based on instance labels and details are shown in the appendix.

\section{Experiments and Discussion}

\begin{table*}[ht]
\centering
\small
\begin{tabular}{@{}clcccccc@{}}
\toprule \midrule
\multicolumn{8}{c}{ZH}                                                                                                                                            \\ \midrule
\#id & \textit{Methods}           & \textit{OCNLI} & \textit{TNEWS} & \textit{AFQMC} & \textit{IFLYTEK}            & \textit{WSC}                & \textit{CSL}   \\ \midrule
1    & MacBERT                    & \textbf{78.71}          & 58.76          & 75.49          & 61.45                       & 87.17                       & 83.97          \\
2    & \textit{\textbackslash{}w} \textit{KNN}      & 78.20(-0.51)   & 58.87(+0.11)   & 75.60(+0.11)   & 61.60(+0.15)                & 86.84(-0.33)                & 83.40(-0.57)   \\
3    & \textit{\textbackslash{}w} \textit{KNN best} & 78.71(+0.00)   & 58.97(+0.21)   & 75.63(+0.14)   & 61.91(+0.46)                & 87.17(+0.00)                & 83.97(+0.00)   \\ \midrule
4    & MacBERT+TL                 & \textit{78.07} & 58.83          & 74.95          & 61.60                       & 89.80                       & 83.87          \\
5    & \textit{\textbackslash{}w} \textit{KNN}      & 77.83(-0.24)   & 58.33(-0.60)   & 75.53(+0.58)   & 62.14(+0.54)                & 90.13(+0.33)                & 83.40(-0.47)   \\
6    & \textit{\textbackslash{}w} \textit{KNN best} & 78.07(+0.00)   & 58.83(+0.00)   & 75.53(+0.58)   & 62.14(+0.54)                & 90.13(+0.33)                & 83.87(+0.00)   \\ \midrule
7    & MacBERT+TL+D               & 78.61 & \textbf{59.18} & \textbf{76.18} & \textbf{61.83}              & \textbf{91.12}              & \textbf{84.10} \\
8    & \textit{\textbackslash{}w} \textit{KNN}      & 78.85(+0.24)   & 59.36(+0.18)   & 75.74(-0.44)   & 61.68(-0.15)                & 90.79(-0.33)                & 84.00(-0.10)   \\
9 &
  \textit{\textbackslash{}w} \textit{KNN best} &
  {\ul \textbf{78.98(+0.37)}} &
  {\ul \textbf{59.37(+0.19)}} &
  {\ul \textbf{76.18(+0.00)}} &
  {\ul \textbf{62.25(+0.42)}} &
  {\ul \textbf{91.12(+0.00)}} &
  {\ul \textbf{84.30(+0.20)}} \\ \midrule \midrule
\multicolumn{8}{c}{EN}                                                                                                                                            \\ \midrule
     & \textit{Methods}           & \textit{CoLA}  & \textit{SST-2} & \textit{MRPC}  & \textit{QQP}                & \textit{MNLI-m/mm}          & \textit{QNLI}  \\ \midrule
10   & RoBERTa                    & 67.49          & 96.44          & 90.69          & \textbf{92.18}              & 90.15                       & 94.76          \\
11   & \textit{\textbackslash{}w} \textit{KNN}      & 67.86(+0.37)   & 96.44(+0.00)   & 90.44(-0.15)   & 92.16(-0.02)                & 90.31(+0.16)                & 94.64(-0.12)   \\
12   & \textit{\textbackslash{}w} \textit{KNN best} & 68.53(+1.04)   & 96.44(+0.00)   & 90.69(+0.00)   & {\ul \textbf{92.19(+0.01)}} & 90.33(+0.18)                & 94.76(+0.00)   \\ \midrule
13   & RoBERTa+TL                 & 67.70          & 96.10          & 90.20          & 92.00                       & \textbf{90.27}              & 94.58          \\
14   & \textit{\textbackslash{}w} \textit{KNN}      & 66.85(-0.15)   & 96.22(+0.12)   & 89.95(-0.25)   & 92.05(+0.05)                & 90.38(+0.11)                & 94.45(-0.13)   \\
15   & \textit{\textbackslash{}w} \textit{KNN best} & 67.70(+0.00)   & 96.22(+0.12)   & 90.20(+0.00)   & 92.05(+0.05)                & {\ul \textbf{90.41(+0.14)}} & 94.58(+0.00)   \\ \midrule
16   & RoBERTa+TL+D               & \textbf{69.06} & \textbf{96.56} & \textbf{90.93} & 92.07                       & 90.33                       & \textbf{95.06} \\
17   & \textit{\textbackslash{}w} \textit{KNN}      & 70.27(+1.11)   & 96.44(-0.12)   & 90.93(+0.00)   & 92.07(+0.00)                & 90.15(-0.18)                & 95.08(+0.02)   \\
18 &
  \textit{\textbackslash{}w} \textit{KNN best} &
  {\ul \textbf{70.49(+1.43)}} &
  {\ul \textbf{96.56(+0.00)}} &
  {\ul \textbf{90.93(+0.00)}} &
  92.08(+0.01) &
  90.33(+0.00) &
  {\ul \textbf{95.15(+0.09)}} \\ \midrule \bottomrule
\end{tabular}
\caption{Results on 12 classification datasets. \textbackslash{}w \textit{KNN}: only use retrieved labels to obtain final label. \textbackslash{}w \textit{KNN best}: interpolated results of prediction and retrieval distributions. TL: training with both classification and triplet loss. D: training with the decouple layer.}
\label{tab:main}
\end{table*}

\subsection{Evaluation Settings}
We evaluate our methods on six Chinese and six English classification datasets, including OCNLI~\cite{hu-etal-2020-ocnli}, TNEWS, AFQMC, IFLYTEK, WSC, CSL, CoLA, SST-2, MRPC, QQP, MNLI-m/mm~\cite{williams-etal-2018-broad}, and QNLI. We separately choose MacBERT-large~\cite{cui-etal-2020-revisiting,cui-etal-2021-pretrain} and RoBERTa-large~\cite{roberta} as our base PLMs for Chinese and English. For each instance during testing, we retrieve 64 nearest neighbours to intervene in the prediction distribution. The hyper-parameter $\beta$ is set as 0.5 and the temperature $T$ is set as 10.
The details of training settings are shown in the appendix.

\begin{table}[ht]
\centering
\small
\begin{tabular}{@{}lccc@{}}
\toprule \midrule
\textit{Vector} & \textit{CLS}   & \textit{MEAN}  & \textit{MAX} \\ \midrule
\multicolumn{4}{c}{ZH}                                  \\ \midrule
OCNLI           & \textbf{78.98} & 77.96          & 77.01        \\
TNEWS           & \textbf{59.37} & 59.17          & 57.89        \\
AFQMC           & 76.18          & \textbf{76.34} & 76.04        \\
IFLYTEK         & \textbf{62.25} & 61.87          & 61.31        \\
WSC             & \textbf{91.12} & \textbf{91.12} & 90.28        \\
CSL             & 84.30          & \textbf{84.57} & 83.96        \\ \midrule \midrule
\multicolumn{4}{c}{EN}                                  \\ \midrule
CoLA            & \textbf{70.49} & 69.24          & 68.79        \\
SST-2           & 95.56          & \textbf{95.99} & 94.98        \\
MRPC            & \textbf{90.93} & 90.69          & 89.84        \\
QQP             & 92.08          & \textbf{92.15} & 91.39        \\
MNLI-m/mm       & 90.50          & \textbf{90.64} & 89.74        \\
QNLI            & \textbf{95.15} & 94.85          & 94.17        \\ 
 \midrule \bottomrule
\end{tabular}
\caption{Results with different retrieval vectors.}
\label{tab:revec}
\end{table}

\subsection{Main Results}
From the results in Table~\ref{tab:main}, we can summarize some conclusions as follows:


1) By comparing the results \#id 1-3,10-12 we can conclude that the KNN-based model can slightly improve the performance of classification tasks. 

2) We can see that results \#id 4-6,13-15 decreased after we add triplet loss (TL) loss to help the model learn instance representations.

3) The overall results show that the combination of decoupled representation (D) and triplet loss significantly improves the classification performance, which proves the decoupling mechanism for learning separate representations is necessary for KNN-based classification models.

4) Interesting, we find that our proposed decoupling mechanism can brings significant gain in results to MacBERT/RoBERTa, e.g. the \#id 7 and 16. However, the improvement from KNN retrieved instances disappears in many datasets. We guess that the modelling of instance representations help the model cluster similar instances in embedding space. Then, the improvement from KNN is offset by this representation learning. We will check this in future work.




\subsection{Discussion}
\paragraph{The Impact of Retrieval Representations}
We replace the retrieval representation with the meaning of all tokens' vectors (MEAN) or the max-pooling of all tokens' vectors (MAX) to investigate what representation is better for KNN-based models.
From Table~\ref{tab:revec}, we can summarize that the CLS vector $h_0$ achieves the overall best performance and is the most stable retrieval vector. The MEAN can achieve better results on some datasets. However, the MAX obtained terrible performance. So for different datasets, the MEAN and CLS vectors can be attempted for getting better results.

\paragraph{The Effect of Hyper-parameters}
\begin{figure}[ht]
    \centering
    \includegraphics[width=0.4\textwidth]{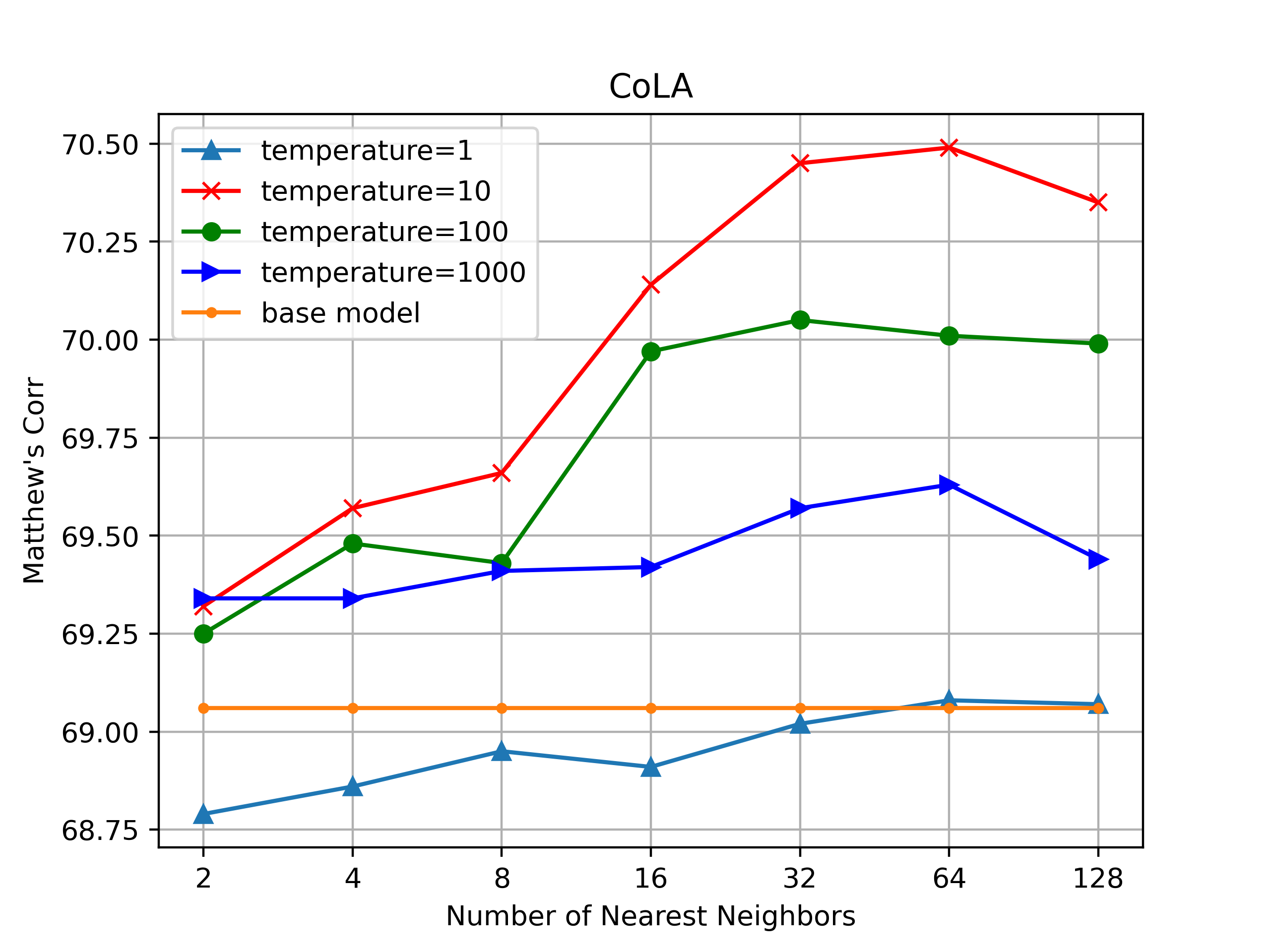}
    \caption{The effect of hyper-parameters $k$ and $T$.}
    \label{fig:tn}
\end{figure} 

We also show the performance of our method with different retrieved neighbours $k$ and scale weight temperatures $T$. We can see that from Figure~\ref{fig:tn}, the model is sensitive to the hyper-parameter temperature and the best setting is 10. For the number of neighbours, we can see that the performance increases with the increase in numbers. However, when $k$ is set as 128, the performance begins to fluctuate. Finally, the best number of neighbours is 64.

\section{Conclusion}
In this paper, we propose a $k$-nearest-neighbor (KNN) -based method for retrieval augmented classifications, which uses the $k$-nearest neighbour model to retrieve information from training data and employs the retrieved $k$ labels to directly interpolate the predicted label distribution. In addition, we also propose a simple yet effective decoupling mechanism to ensure KNN-based methods work on classification tasks. Experimental results demonstrate that our proposed methods can effectively improve the performance of a wide range of classification tasks. In future work, we will try to extend this method to more complex NLP tasks, e.g. question answering, and named entity recognition. 

\section*{Limitations}
There are still some limitations of this paper: 1) The motivation of our work is to extend the KNN-based method to more NLP tasks, in this paper, we only extend it to the classification tasks. 2) We only choose two PLMs as our base models and the KNN-based method with more different PLMs should be tried. 3) There are still some special experimental results that can not be interpreted well. We will try to solve these limitations in future work.


\bibliography{custom}
\bibliographystyle{acl_natbib}

\appendix

\section{Training Settings}
\begin{table*}[]
\centering
\small
\begin{tabular}{lcccccc}
\toprule \midrule 
\multicolumn{7}{c}{\textbf{ZH}}                                                           \\ \midrule 
\textit{config}  & \textit{OCNLI} & \textit{TNEWS} & \textit{AFQMC} & \textit{IFLYTEK} & \textit{WSC}       & \textit{CSL}  \\ \midrule
\textbf{Best Lambda}     & 0.2     & 0.8     & 0.0     & 0.1          & 0.1     & 0.2     \\
\textbf{Batch Size}      & 64      & 64      & 64      & 64           & 64      & 64      \\
\textbf{Learning Rate}   & 2E-05   & 2E-05   & 2E-05   & 2E-05        & 2E-05   & 2E-05   \\
\textbf{Init Model}      & MacBERT & MacBERT & MacBERT & MacBERT      & MacBERT & MacBERT \\
\textbf{Warmup Steps}    & 100     & 300     & 100     & 100          & 300     & 300     \\
\textbf{Training Epochs} & 3       & 5       & 3       & 3            & 3       & 3       \\
\textbf{Max Len} & 150   & 200   & 128   & 512     & 200       & 512  \\ \midrule  \midrule
\multicolumn{7}{c}{\textbf{EN}}                                                           \\ \midrule
\textit{config}  & \textit{CoLA}  & \textit{SST-2} & \textit{MRPC}  & \textit{QQP}     & \textit{MNLI-m/mm} & \textit{QNLI} \\ \midrule 
\textbf{Best Lambda}     & 0.3     & 0.0     & 0.0     & 0.5 & 0.0     & 0.2     \\
\textbf{Batch Size}      & 32      & 32      & 32      & 32           & 32      & 32      \\
\textbf{Learning Rate}   & 1E-5    & 1E-5    & 1E-5    & 1E-5         & 1E-5    & 1E-5    \\
\textbf{Init Model}      & RoBERTa & RoBERTa & RoBERTa & RoBERTa      & RoBERTa & RoBERTa \\
\textbf{Warmup Steps}    & 100     & 100     & 100     & 100          & 100     & 100     \\
\textbf{Training Epochs} & 3       & 3       & 3       & 3            & 3       & 3       \\
\textbf{Max Len}         & 128     & 128     & 128     & 128          & 128     & 128     \\ \midrule \bottomrule
\end{tabular}
\caption{The fine-tuning setting of our experiments.}
\label{tab:hyparam}
\end{table*}

We show the settings of all datasets in Table~\ref{tab:hyparam}. The default $\beta$ is 0.5, nearest neighbors $k$ is 64, and temperature $T$ is 10. 

\section{Positive\&Negative Examples Selection}
In this section, we introduce the details to select positive and negative examples for learning instance representations.
Specifically, we select instances with the same label in the same batch as positive examples and instances with a different label as negative examples. If the positive example does not exist, we directly employ the instance itself as the positive example. If the negative example does not exist, we randomly sample any other example as the negative example.





\end{document}